\documentclass[conference]{IEEEtran}
\usepackage{times}

\usepackage[bookmarks=true]{hyperref}
\usepackage{multicol}
\usepackage[numbers]{natbib}
\usepackage{booktabs}
\usepackage{makecell}
\usepackage{tikz}
\usetikzlibrary{shapes,arrows.meta, positioning, calc, fit, backgrounds}

\newcommand{\name}{$SLowRL$ }

\newcommand{\x}{\mathbf{x}}

\newcommand{\y}{\mathbf{y}}

\usepackage{times}
 \usepackage[printonlyused]{acronym}
\usepackage[export]{adjustbox} 
\usepackage{algorithm}
\usepackage{algorithmicx}
\usepackage{algpseudocode}
\algrenewcommand\algorithmicindent{0.8em}%
\usepackage{amssymb}
\usepackage{amsmath}
\usepackage{array}
\usepackage{bbm}
\usepackage{booktabs}
\usepackage{caption}
\usepackage{float}
\usepackage{graphicx}
\usepackage{cleveref}
\usepackage{lipsum}
\usepackage{microtype}
\usepackage{multicol}
\usepackage{multirow}
\usepackage{nicefrac}
\usepackage{relsize}
\usepackage{siunitx}
\usepackage{subcaption}
\usepackage{todonotes}
\usepackage{wrapfig}
\usepackage{tikz}
\usepackage{verbatim}
\usepackage{xspace}

\usepackage{amsthm}
\algnewcommand{\IfThenElse}[3]{
  \State \algorithmicif\ #1\ \algorithmicthen\ #2\ \algorithmicelse\ #3}
 \algnewcommand{\IfThen}[2]{
  \State \algorithmicif\ #1\ \algorithmicthen\ #2}

\usepackage[utf8]{inputenc} 
\usepackage[T1]{fontenc}    
\usepackage{url}            
\usepackage{booktabs}       
\usepackage{amsfonts}       
\usepackage{nicefrac}       
\usepackage{microtype}      

\definecolor{mydarkblue}{rgb}{0,0.08,0.45}
\hypersetup{ %
    pdftitle={},
    pdfauthor={},
    pdfsubject={},
    pdfkeywords={},
    pdfborder=0 0 0,
    pdfpagemode=UseNone,
    colorlinks=true,
    linkcolor=mydarkblue,
    citecolor=mydarkblue,
    filecolor=mydarkblue,
    urlcolor=mydarkblue,
    pdfview=FitH}

\algnewcommand{\algorithmicforeach}{\textbf{for each}}
\algdef{SE}[FOR]{ForEach}{EndForEach}[1]
  {\algorithmicforeach\ #1\ \algorithmicdo}
  {\algorithmicend\ \algorithmicforeach}
\newcommand{\pushright}[1]{\ifmeasuring@#1\else\omit\hfill$\displaystyle#1$\fi\ignorespaces}

\usetikzlibrary{tikzmark}
\usetikzlibrary{calc}
\newcommand{\ALGtikzmarkcolor}{black}
\newcommand{\ALGtikzmarkextraindent}{2pt}
\newcommand{\ALGtikzmarkverticaloffsetstart}{-.5ex}
\newcommand{\ALGtikzmarkverticaloffsetend}{-.5ex}
\makeatletter
\newcounter{ALG@tikzmark@tempcnta}

\newcommand\ALG@tikzmark@start{%
    \global\let\ALG@tikzmark@last\ALG@tikzmark@starttext%
    \expandafter\edef\csname ALG@tikzmark@\theALG@nested\endcsname{\theALG@tikzmark@tempcnta}%
    \tikzmark{ALG@tikzmark@start@\csname ALG@tikzmark@\theALG@nested\endcsname}%
    \addtocounter{ALG@tikzmark@tempcnta}{1}%
}

\def\ALG@tikzmark@starttext{start}
\newcommand\ALG@tikzmark@end{%
    \ifx\ALG@tikzmark@last\ALG@tikzmark@starttext
    \else
        \tikzmark{ALG@tikzmark@end@\csname ALG@tikzmark@\theALG@nested\endcsname}%
        \tikz[overlay,remember picture] \draw[\ALGtikzmarkcolor] let \p{S}=($(pic cs:ALG@tikzmark@start@\csname ALG@tikzmark@\theALG@nested\endcsname)+(\ALGtikzmarkextraindent,\ALGtikzmarkverticaloffsetstart)$), \p{E}=($(pic cs:ALG@tikzmark@end@\csname ALG@tikzmark@\theALG@nested\endcsname)+(\ALGtikzmarkextraindent,\ALGtikzmarkverticaloffsetend)$) in (\x{S},\y{S})--(\x{S},\y{E});%
    \fi
    \gdef\ALG@tikzmark@last{end}%
}

\apptocmd{\ALG@beginblock}{\ALG@tikzmark@start}{}{\errmessage{failed to patch}}
\pretocmd{\ALG@endblock}{\ALG@tikzmark@end}{}{\errmessage{failed to patch}}

\begin{document}

\title{SLowRL: \underline{S}afe \underline{Low}-Rank Adaptation \underline{R}einforcement \underline{L}earning for Locomotion}

\author{
    \IEEEauthorblockN{
        Elham Daneshmand\IEEEauthorrefmark{1}\IEEEauthorrefmark{2}, 
        Shafeef Omar\IEEEauthorrefmark{4}, 
        Glen Berseth\IEEEauthorrefmark{2}\IEEEauthorrefmark{3}, 
        Majid Khadiv\IEEEauthorrefmark{4}, 
        Hsiu-Chin Lin\IEEEauthorrefmark{1}
    }
    \IEEEauthorblockA{
        \IEEEauthorrefmark{1}McGill University, 
        \IEEEauthorrefmark{2}Mila - Quebec AI Institute, 
        \IEEEauthorrefmark{3}Université de Montréal, 
        \IEEEauthorrefmark{4}Technical University of Munich
    }
}


\acrodef{AGI}{artificial general intelligence}
\acrodef{ANOVA}[ANOVA]{Analysis of Variance\acroextra{, a set of
  statistical techniques to identify sources of variability between groups}}
\acrodef{ANN}{artificial neural network}
\acrodef{API}{application programming interface}
\acrodef{CACLA}{continuous actor critic learning automaton}
\acrodef{cGAN}{conditional generative adversarial network}
\acrodef{CMA}{covariance matrix adaptation}
\acrodef{COM}{centre of mass}
\acrodef{CTAN}{\acroextra{The }Common \TeX\ Archive Network}
\acrodef{DDPG}{deep deterministic policy gradient}
\acrodef{DeepLoco}{deep locomotion}
\acrodef{DOI}{Document Object Identifier\acroextra{ (see
    \url{http://doi.org})}}
\acrodef{DPG}{deterministic policy gradient}
\acrodef{DQN}{deep Q-network}
\acrodef{DRL}{deep reinforcement learning}
\acrodef{DYNA}{DYNA}
\acrodef{EOM}{Equations of motion}
\acrodef{EPG}{expected policy gradient}
\acrodef{FDR}{future discounted reward}
\acrodef{FSM}{finite state machine}
\acrodef{GAE}{generalized advantage estimation}
\acrodef{GAIfO}{generative adversarial imitation from observation}
\acrodef{GAN}{generative adversarial network}
\acrodef{GPS}[GPS]{Graduate and Postdoctoral Studies}
\acrodef{HLC}{high-level controller}
\acrodef{HLP}{high-level policy}
\acrodef{HRL}{hierarchical reinforcement learning}
\acrodef{KLD}{Kullback-Leibler divergence}
\acrodef{LLC}{low-level controller}
\acrodef{LLP}{low-level policy}
\acrodef{MARL}{Multi-Agent Reinforcement Learning}
\acrodef{MBAE}{model-based action exploration}
\acrodef{MPC}{model predictive control}
\acrodef{MDP}{Markov Decision Processes}
\acrodef{MSE}{mean squared error}
\acrodef{MultiTasker}{controller that learns multiple tasks at the same time}
\acrodef{Parallel}{randomly initialize controllers and train in parallel}
\acrodef{PD}{proportional derivative}
\acrodef{PDF}{Portable Document Format}
\acrodef{PLAiD}{Progressive Learning and Integration via Distillation}
\acrodef{PPO}{proximal policy optimization}
\acrodef{PTD}{positive temporal difference}
\acrodef{RBF}{radial basis function}
\acrodef{ReLU}{rectified linear unit}
\acrodef{RCS}[RCS]{Revision control system\acroextra{, a software
    tool for tracking changes to a set of files}}
\acrodef{RL}{reinforcement learning}
\acrodef{SGD}{stochastic gradient descent}
\acrodef{Scratch}{randomly initialized controller}
\acrodef{SIMBICON}{SIMple BIped CONtroller}
\acrodef{SMBAE}{stochastic model-based action exploration}
\acrodef{SVG}{stochastic value gradients}
\acrodef{SVM}{support vector machine}
\acrodef{TCN}{time contrastive learning}
\acrodef{TD}{temporal difference}
\acrodef{TL}{transfer learning}
\acrodef{terrainRL}{terrain adaptive locomotion}
\acrodef{TLX}[TLX]{Task Load Index\acroextra{, an instrument for gauging
  the subjective mental workload experienced by a human in performing
  a task}}
\acrodef{TRPO}{trust region policy optimization}
\acrodef{UBC}{University of British Columbia}
\acrodef{UCB}{upper confidence bound}
\acrodef{UI}{user interface}
\acrodef{UML}{Unified Modelling Language\acroextra{, a visual language
    for modelling the structure of software artefacts}}
\acrodef{URDF}{unified robot description format}
\acrodef{URL}{Unique Resource Locator\acroextra{, used to describe a
    means for obtaining some resource on the world wide web}}
\acrodef{W3C}[W3C]{\acroextra{the }World Wide Web Consortium\acroextra{,
    the standards body for web technologies}}    
\acrodef{XML}{Extensible Markup Language}
\acrodef{DR}{domain randomization}
\acrodef{LoRA}{Low-Rank Adaptation}
\acrodef{PEFT}{Parameter-Efficient Fine-Tuning}
\acrodef{FFT}{Full Fine-Tuning}

\maketitle

\setlength{\abovecaptionskip}{3pt}
\setlength{\belowcaptionskip}{0pt}

\begin{abstract}
Sim-to-real transfer of locomotion policies often leads to performance degradation due to the inevitable sim-to-real gap. Naively fine-tuning these policies directly on hardware is problematic, as it poses risks of mechanical failure and suffers from high sample inefficiency.
In this paper, we address the challenge of safely and efficiently fine-tuning reinforcement learning (RL) policies for dynamic locomotion tasks. Specifically, we focus on fine-tuning policies learned in simulation directly on hardware, while explicitly enforcing safety constraints. In doing so, we introduce \name—a framework that combines Low-Rank Adaptation (LoRA) with training-time safety enforcement via a recovery policy. We evaluate our method both in simulation and on a real Unitree Go2 quadruped robot for jump and trot tasks. Experimental results show that our method achieves a $46.5\%$ reduction in fine-tuning time and near-zero safety violations compared to standard proximal policy optimization (PPO) baselines. Notably, we find that a rank-1 adaptation alone is sufficient to recover pre-trained performance in the real world, while maintaining stable and safe real-world fine-tuning. These results demonstrate the practicality of safe, efficient fine-tuning for dynamic real-world robotic applications.
\end{abstract}

\section{Introduction}
\label{sec:intro}
Achieving high performance on real-world robotic systems remains a fundamental challenge in robotics due to hardware constraints and the persistent sim-to-real gap. The ideal pipeline involves leveraging massively parallelized simulation to explore complex environments and learn robust behaviors, followed by a transition to physical platforms for targeted real-world fine-tuning. The goal of this final stage is to achieve peak performance by adapting to the specific nuances of the robot and the world; however, executing this adaptation without damaging the robot remains a significant hurdle. 


\begin{figure}[h] 
    \centering
    \resizebox{\linewidth}{!}{\input{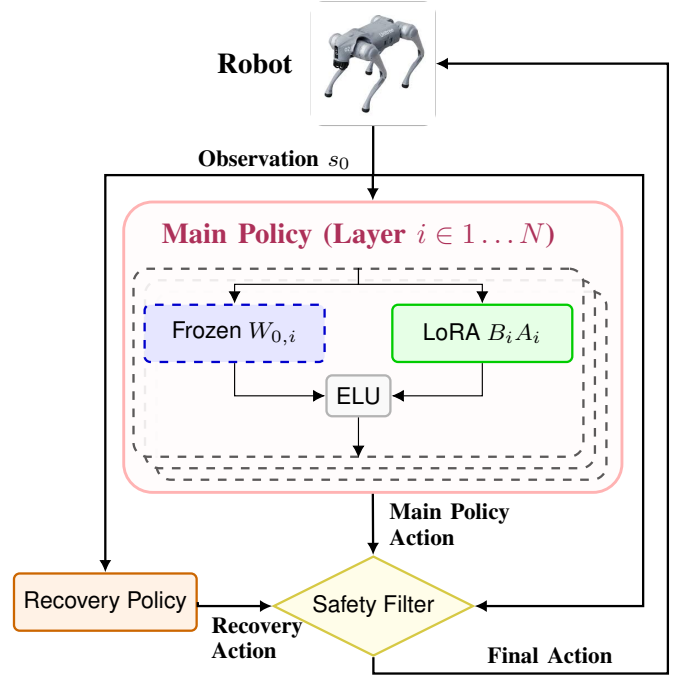}} 
    \caption{The \name architecture. A frozen main policy and a trainable LoRA adapter operate in parallel to generate the main action. A Safety Filter monitors the robot's state and selects between the main action and a conservative Recovery Policy action to ensure safe operation during fine-tuning.}
    \label{fig:1}
    \vspace{-0.45cm}
\end{figure}

Learning-based approaches, particularly deep \ac{RL}, have demonstrated impressive performance and versatility across a wide range of locomotion behaviors \citep{ha2025learning}. 
Nevertheless, transfer to the real world often causes a certain degree of performance degradation, largely due to distributional mismatch between simulation and hardware. Reported results indicate that full-model fine-tuning requires hours of prohibitive real-world interaction, creating a significant bottleneck for scalable deployment and risking hardware damage during exploration \citep{smith2021leggedrobotslearningfinetuning, thananjeyan2021recoveryrlsafereinforcement, yang2022safereinforcementlearninglegged}. As a result, there remains a lack of unified frameworks that can safely and efficiently improve policies in the real world. 


In this work, we hypothesize that effective sim-to-real adaptation does not require retraining the entire policy, but can be achieved by optimizing a low-rank subspace of parameters. Our empirical results reveal that this adaptation can be achieved with extremely low-rank updates, with rank-1 adaptation consistently providing the fastest recovery of pre-trained performance. To test this hypothesis in a safety-critical real-world setting, we introduce \name—a framework that combines low-rank policy adaptation with online safety enforcement. This combination of parameter-efficient fine-tuning and a learned safety filter enables us to minimize the duration of real-robot interactions while preventing hardware failures. Our experimental results demonstrate that this approach achieves a $46.5\%$ reduction in fine-tuning time with near-zero safety violations compared to standard \ac{FFT} using \ac{PPO} baselines.

Unlike prior methods that treat real-world learning time as a secondary concern, we explicitly focus on minimizing the duration of real-robot fine-tuning with minimal falls,
while achieving comparable or improved task performance. We validate our framework across multiple locomotion tasks using a single set of hyperparameters and demonstrate consistent improvements in fine-tuning efficiency. Our results show more improvement during 60 minutes of training in the real-world compared to \ac{FFT} baselines.

In summary, this work argues that sim-to-real adaptation for legged locomotion is fundamentally a low-dimensional policy refinement problem. We show that effective real-world fine-tuning can be achieved through extremely low-rank policy updates, with a single rank-1 adaptation direction sufficient to recover and improve performance relative to \ac{FFT}. We further demonstrate that stable real-world adaptation requires jointly adapting both the actor and critic, and that incorporating explicit safety mechanisms enables reliable real-world fine-tuning with substantially reduced failure rates and faster empirical convergence. These claims are validated through extensive sim-to-sim and sim-to-real experiments on a quadruped robot, including dynamic locomotion behaviors, demonstrating that safe and efficient real-world adaptation does not require full policy retraining.

\section{Related Work}
\label{sec:intro}

Deep RL, specifically \ac{PPO} \citep{schulman2017proximalpolicyoptimizationalgorithms}, has become extensively utilized for the synthesis of legged locomotion controllers \citep{ha2025learning}. Despite improvements in the quality of simulation environments, the sim-to-real gap remains a critical bottleneck to achieving multipurpose robots. Nevertheless, \ac{DR} has proven to be effective in successfully transferring policies to the real world \citep{8202133, tan2018simtoreallearningagilelocomotion, Peng_2018,bogdanovic2022modelfreereinforcementlearningrobust}. This comes at the cost of overly conservative policies that fail to exploit the robot's full dynamic potential. Alternatives such as explicit system identification \citep{Hwangbo_2019,10.1109/ICRA.2019.8793789} or latent-space adaptation \citep{kumar2021rmarapidmotoradaptation, doi:10.1126/scirobotics.abk2822} have shown promise to circumvent this issue. However, these methods generally assume that the real world can be represented as in-distribution relative to ensembles of simulations, which is a strong and often unrealistic assumption.

To address out-of-distribution scenarios, recent works have explored fine-tuning policies directly on hardware. A primary challenge in this domain is safety. Approaches like Recovery RL \citep{thananjeyan2021recoveryrlsafereinforcement} or safety-constrained frameworks \citep{liu2023constrained} utilize safety critics or constraint layers to intervene during dangerous states. However, relying on a critic or constraint model trained in simulation can be perilous, as the safety signal itself is subject to the sim-to-real gap. 
Furthermore, existing fine-tuning frameworks suffer from severe sample inefficiency. For instance, \citep{yang2022safereinforcementlearninglegged} reports adaptation times of approximately 115 minutes per task—a prohibitive duration for scaling to diverse behaviors. While \citep{smith2023growlimitscontinuousimprovement} achieves faster learning (approx. 20 minutes), it relies on the assumption that the robot is statically stable within a restricted action space. This assumption renders the method inapplicable to underactuated platforms such as underactuated bipedal robots \citep{Daneshmand_2021}, which require continuous active balance.



We bridge these gaps by introducing \name, a framework that enables rapid, safe adaptation on real robots. In particular, we leverage \ac{LoRA} \citep{hu2022lora, ding2023parameter}, a fine-tuning paradigm used in large language models (LLMs) to address similar inefficiencies by optimizing only a low-dimensional subspace. By restricting optimization to a low-rank manifold, we significantly reduce sample complexity and mitigate the safety risks associated with prolonged on-hardware fine-tuning. Furthermore, updating only a low-rank subspace prevents the bang-bang control spikes often seen in over-parameterized models. To the best of our knowledge, the application of \ac{PEFT} techniques \cite{han2024parameter} to high-frequency robotic control has not been explored. On top of using LoRA, we explicitly integrate a learned recovery policy during fine-tuning of the policies, preventing the exploration of dangerous parts of the state space.

\section{Background}
\label{sec:background}

\subsection{Reinforcement Learning Formulation}
We formulate the robot control problem as a \ac{MDP} defined by the tuple $(\mathcal{S}, \mathcal{A}, \mathcal{P}, \mathcal{R}, \gamma)$. At each time step $t$, the robot is in state $s_t \in \mathcal{S} \subseteq \mathbb{R}^n$ and executes an action $a_t \in \mathcal{A} \subseteq \mathbb{R}^m$ according to a stochastic policy $\pi_\theta(a_t | s_t)$ parameterized by weights $\theta$. The environment dynamics are governed by the transition probability $\mathcal{P}(s_{t+1} | s_t, a_t)$. The goal of the \ac{RL} agent is to learn optimal parameters $\theta^*$ that maximize the expected discounted cumulative reward:
$$J(\pi_\theta) = \mathbb{E}_{\tau \sim \pi_\theta} \left[ \sum_{t=0}^{T} \gamma^t r_t \right]$$where $r_t = \mathcal{R}(s_t, a_t)$ is the reward at time $t$, $T$ is the horizon, and $\gamma \in [0, 1)$ is the discount factor. In this work, we utilize \ac{PPO} \citep{schulman2017proximalpolicyoptimizationalgorithms}, a widely adopted algorithm for legged locomotion control \citep{rudin2022learningwalkminutesusing, tan2018simtoreallearningagilelocomotion, doi:10.1126/scirobotics.abk2822, peng2020learningagileroboticlocomotion}, due to its stability and sample efficiency. In the remainder of the paper, we instantiate $\theta$ using neural network weights $W$, with $W_0$ denoting the frozen pre-trained parameters.

\subsection{Low-Rank Adaptation (LoRA)}\label{LoRA}
Standard fine-tuning typically involves updating the entire pre-trained weight matrix $W_0 \in \mathbb{R}^{d \times k}$ to adapt to a new distribution. When referring to individual layers, we denote the corresponding frozen weights as $W_{0,i}$; for simplicity, we use $W_0$ to represent the collection of all frozen policy parameters. The adapted weights can be expressed as $W = W_0 + \Delta W$, where $\Delta W$ represents the environment-specific perturbation required to align the policy with the physical dynamics of the real robot.

Low-Rank Adaptation (LoRA)  \citep{hu2022lora} is a parameter-efficient fine-tuning method originally introduced for adapting large neural networks—especially large language models without updating all of their parameters.
Low-Rank Adaptation (LoRA) hypothesizes that the change in weights $\Delta W$ during adaptation resides in a low intrinsic rank $\rho \ll \min(d, k)$. \ac{LoRA} decomposes the update into two low-rank matrices $B \in \mathbb{R}^{d \times \rho}$ and $A \in \mathbb{R}^{\rho \times k}$. The forward pass for a linear layer is reformulated as:
\begin{equation}
    h = W_0 x + \Delta W x = W_0 x + B A x
    \label{equ:LoRA}
\end{equation}

\noindent
During policy refinement, $W_0$ is frozen, and $B A$ are refinable parameters.
LoRA has proven effective for refining large language models; in this work, we repurpose it for sim-to-real policy refinement.

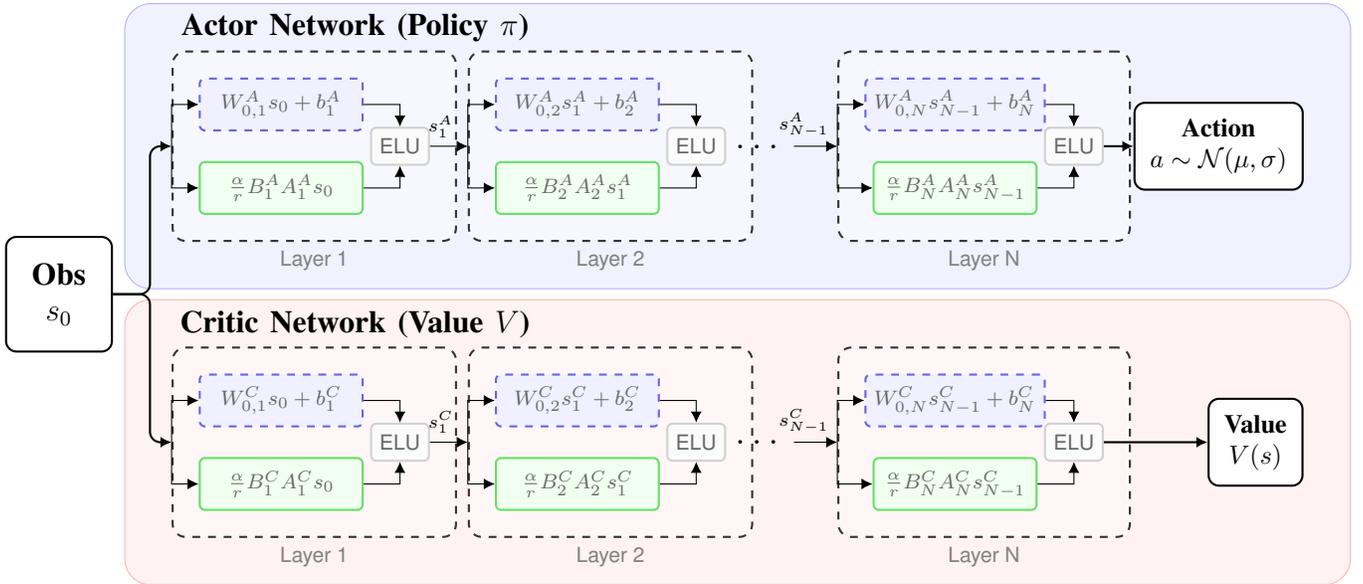
\begin{figure*}[t]
    \centering
    \resizebox{\linewidth}{!}{\usetikzlibrary{shapes, arrows.meta, positioning, calc, fit, backgrounds}

\begin{tikzpicture}[
    node distance=1.2cm and 1.5cm,
    font=\sffamily\footnotesize,
    >={Latex[width=1.5mm,length=1.5mm]},
    frozen/.style={
        rectangle, draw=blue!80!black, fill=blue!10, dashed, thick, 
        align=center, minimum height=0.7cm, minimum width=2.2cm, rounded corners=2pt
    },
    trainable/.style={
        rectangle, draw=green!80!black, fill=green!10, thick, 
        align=center, minimum height=0.7cm, minimum width=2.2cm, rounded corners=2pt
    },
    activation/.style={
        rectangle, draw=gray!60, fill=gray!5, thick, 
        align=center, minimum size=0.5cm, rounded corners=2pt
    },
    layerbox/.style={
        draw=black!80, dashed, thick, inner sep=10pt, rounded corners=5pt, fill=white, fill opacity=0.4
    }
]

    \node [font=\large\bfseries, align=center, draw, thick, rounded corners, inner sep=10pt, fill=white] (Obs) at (0,0) {Obs\\$s_0$};

    \begin{scope}[yshift=2.0cm, xshift=3cm]
        \node [font=\large\bfseries] (TitleA) at (1.0, 1.6) {Actor Network (Policy $\pi$)};
        
        \coordinate (A1Start) at (0,0);
        \node [frozen, above=0.2cm of A1Start] (w1A) {$W^A_{0,1} s_0 + b^A_1$};
        \node [trainable, below=0.2cm of A1Start] (lora1A) {$\frac{\alpha}{r} B^A_1 A^A_1 s_0$};
        \node [activation, right=1.2cm of A1Start] (act1A) {ELU};
        \node [layerbox, fit=(w1A) (lora1A) (act1A), label={[gray]below:Layer 1}] (L1BoxA) {};

        \draw [->] (w1A.east) -| (act1A.north);
        \draw [->] (lora1A.east) -| (act1A.south);
        \draw [->] (L1BoxA.west) -- ++(-0.0,0.5) |- (w1A.west);
        \draw [->] (L1BoxA.west) -- ++(-0.0,-0.5) |- (lora1A.west);
        \coordinate (A2Start) at (4.0,0);
        \node [frozen, above=0.2cm of A2Start] (w2A) {$W^A_{0,2} s^A_1 + b^A_2$};
        \node [trainable, below=0.2cm of A2Start] (lora2A) {$\frac{\alpha}{r} B^A_2 A^A_2 s^A_1$};
        \node [activation, right=1.2cm of A2Start] (act2A) {ELU};
        \node [layerbox, fit=(w2A) (lora2A) (act2A), label={[gray]below:Layer 2}] (L2BoxA) {};
        
        \draw [->] (act1A) -- node[above right=0.0cm and -0.4cm, font=\scriptsize] {$s^A_1$} (L2BoxA);
        \draw [->] (w2A.east) -| (act2A.north);
        \draw [->] (lora2A.east) -| (act2A.south);

        \draw [->] (L2BoxA.west) -- ++(-0.0,0.5) |- (w2A.west);
        \draw [->] (L2BoxA.west) -- ++(-0.0,-0.5) |- (lora2A.west);
        \node [right=0.0cm of act2A, font=\Large] (dotsA) {$\dots$};

        \coordinate (ANStart) at (9.1,0);
        \node [frozen, above=0.2cm of ANStart] (wNA) {$W^A_{0,N} s^A_{N-1} + b^A_N$};
        \node [trainable, below=0.2cm of ANStart] (loraNA) {$\frac{\alpha}{r} B^A_N A^A_N s^A_{N-1}$};
        \node [activation, right=1.2cm of ANStart] (actNA) {ELU};
        \node [layerbox, fit=(wNA) (loraNA) (actNA), label={[gray]below:Layer N}] (LNBoxA) {};
        
        \draw [->] (dotsA) -- node[above left= 0.0cm and -0.3cm,, font=\scriptsize] {$s^A_{N-1}$} (LNBoxA.west);
        \draw [->] (wNA.east) -| (actNA.north);
        \draw [->] (loraNA.east) -| (actNA.south);
        \draw [->] (LNBoxA.west) -- ++(-0.0,0.5) |- (wNA.west);
        \draw [->] (LNBoxA.west) -- ++(-0.0,-0.5) |- (loraNA.west);
        \node [right=0.4cm of actNA, font=\bfseries, draw, thick, rounded corners, inner sep=6pt, fill=white, align=center] (outputA) {Action \\ $a \sim \mathcal{N}(\mu, \sigma)$};
        \draw [->, thick] (actNA) -- (outputA);

        \begin{scope}[on background layer]
            \node [draw=blue!30, fill=blue!5, fit=(L1BoxA) (outputA) (lora1A), rounded corners=10pt, inner sep=18pt] (ActorBG) {};
        \end{scope}
    \end{scope}

    \begin{scope}[yshift=-2.0cm, xshift=3cm]
        \node [font=\large\bfseries] (TitleC) at (1.0, 1.6) {Critic Network (Value $V$)};
        
        \coordinate (C1Start) at (0,0);
        \node [frozen, above=0.2cm of C1Start] (w1C) {$W^C_{0,1} s_0 + b^C_1$};
        \node [trainable, below=0.2cm of C1Start] (lora1C) {$\frac{\alpha}{r} B^C_1 A^C_1 s_0$};
        \node [activation, right=1.2cm of C1Start] (act1C) {ELU};
        \node [layerbox, fit=(w1C) (lora1C) (act1C) , label={[gray]below:Layer 1}] (L1BoxC) {};

        \draw [->] (L1BoxC.west) -- ++(-0.0,0.5) |- (w1C.west);
        \draw [->] (L1BoxC.west) -- ++(-0.0,-0.5) |- (lora1C.west);
        \draw [->] (w1C.east) -| (act1C.north);
        \draw [->] (lora1C.east) -| (act1C.south);

        \coordinate (C2Start) at (4.0,0);
        \node [frozen, above=0.2cm of C2Start] (w2C) {$W^C_{0,2} s^C_1 + b^C_2$};
        \node [trainable, below=0.2cm of C2Start] (lora2C) {$\frac{\alpha}{r} B^C_2 A^C_2 s^C_1$};
        \node [activation, right=1.2cm of C2Start] (act2C) {ELU};
        \node [layerbox, fit=(w2C) (lora2C) (act2C), label={[gray]below:Layer 2}] (L2BoxC) {};
        
        \draw [->] (act1C) -- node[above right= 0.0cm and -0.4cm, font=\scriptsize] {$s^C_1$} (L2BoxC.west);
        \draw [->] (w2C.east) -| (act2C.north);
        \draw [->] (lora2C.east) -| (act2C.south);
        
        \draw [->] (L2BoxC.west) -- ++(-0.0,0.5) |- (w2C.west);
        \draw [->] (L2BoxC.west) -- ++(-0.0,-0.5) |- (lora2C.west);

        \node [right=0.0cm of act2C, font=\Large] (dotsC) {$\dots$};

        \coordinate (CNStart) at (9.1,0);
        \node [frozen, above=0.2cm of CNStart] (wNC) {$W^C_{0,N} s^C_{N-1} + b^C_N$};
        \node [trainable, below=0.2cm of CNStart] (loraNC) {$\frac{\alpha}{r} B^C_N A^C_N s^C_{N-1}$};
        \node [activation, right=1.2cm of CNStart] (actNC) {ELU};
        \node [layerbox, fit=(wNC) (loraNC) (actNC), label={[gray]below:Layer N}] (LNBoxC) {};
        
        \draw [->] (dotsC) -- node[above left= 0.0cm and -0.3cm, font=\scriptsize] {$s^C_{N-1}$} (LNBoxC.west);
        \draw [->] (wNC.east) -| (actNC.north);
        \draw [->] (loraNC.east) -| (actNC.south);

        \draw [->] (LNBoxC.west) -- ++(-0.0,0.5) |- (wNC.west);
        \draw [->] (LNBoxC.west) -- ++(-0.0,-0.5) |- (loraNC.west);
        \node [right=1.4cm of actNC, font=\bfseries, draw, thick, rounded corners, inner sep=6pt, fill=white, align=center] (outputC) {Value \\ $V(s)$};
        \draw [->, thick] (actNC) -- (outputC);

        \begin{scope}[on background layer]
            \node [draw=red!30, fill=red!5, fit=(L1BoxC) (outputC) (lora1C), rounded corners=10pt, inner sep=18pt] (CriticBG) {};
        \end{scope}
        
    \end{scope}

    \draw [->, thick, rounded corners] (Obs.east) -- ++(0.5,0) |- (L1BoxC.west);
    \draw [->, thick, rounded corners] (Obs.east) -- ++(0.5,0) |- (L1BoxA.west);

\end{tikzpicture}} 
    \caption{Detailed schematic of the \name framework. We freeze the pre-trained policy parameters (blue dashed blocks) to retain prior knowledge from IsaacLab. To enable adaptation to the target environment, we inject low-rank trainable adapters (green blocks) in parallel to the frozen weights. The outputs are summed before the ELU activation, ensuring a safe exploration process.
    }
    \label{fig:architecture}
    \vspace{-0.25cm}
\end{figure*}

\section{\name: Efficent Real-World Adaptation}
\label{sec:method} 

\subsection{System Overview}
As illustrated in the conceptual diagram in Fig. \ref{fig:1}, \name operates via a two-stage pipeline:

\textbf{Stage 1: High-Fidelity Pre-training} 
We assume we have a base policy $\pi_\theta$ trained in a high-fidelity simulator. We adopt the contact-explicit architecture from \citep{omar2025learningactcontactunified}, which conditions the agent on desired contact goals (e.g., foot placement locations and timings) rather than implicit task rewards. This enables the main policy $\pi_\theta$ to learn a robust, unified representation of contact dynamics across diverse gaits.

\textbf{Stage 2: Safe Low-Rank Adaptation} Once the base policy  $\pi_\theta$ converges in the source domain, our goal is to refine the policy in the target domain, and we adapted LoRA for the adaptation.
As shown in Figure \ref{fig:1}, our framework consists of the following key components:

\begin{itemize}
\item \textbf{Frozen Base Policy (Blue):} We freeze the weights ($W_0$) of the pre-trained policy from Stage 1 to preserve the foundational locomotion priors. This ensures the robot retains its fundamental walking capabilities throughout the fine-tuning process.
\item \textbf{LoRA Adapter (Green):} We use LoRA modules to capture the environment-specific dynamics. These adapters are applied to individual linear layers within the network (as detailed in Section \ref{LoRA_PPO}), effectively bridging the source-to-target domain gap by optimizing only the low-rank matrices.
\item \textbf{Recovery Policy (Orange):} Since the primary objective is hardware safety, we also include a task-agnostic policy that brings the robots from any state to a pre-defined, safe, nominal state. (Section \ref{sec:safety_mechanisms})

\end{itemize}

\subsection{LoRA-PPO Configuration}
\label{LoRA_PPO}
Recent theoretical work has suggested that reinforcement learning operates in an information-limited regime, in which only a small number of effective update directions can be reliably identified from on-policy data. In particular, \ac{LoRA} Without Regret \citep{schulman2025lora} argues that low-rank—and even rank-1—updates may be sufficient for learning under idealized assumptions.

We use LoRA for refining the policy $\pi_\theta$  in the target domain. In the context of sim-to-real transfer, we interpret $W_0$ as the foundational motor skill acquired in simulation, and the low-rank term $BAx$ from Equation~\ref{equ:LoRA} as the environment-specific perturbation required to align the policy with the physical dynamics of the real robot.

In particular, we freeze the pre-trained weights $W_0$ and only optimize the matrices $A$ and $B$. We initialize $A \sim \mathcal{N}(0, \sigma^2)$ and $B = 0$, ensuring the policy starts exactly at the pre-trained behavior ($BA=0$).
Crucially, this formulation is modular. \ac{LoRA} can be selectively applied to any subset of dense layers within the network architecture, including specific layers of the Actor, or Actor and Critic. 

Figure \ref{fig:architecture} details the internal structure of this integration. The architecture is divided into the Critic (Value $V$) and the Actor (Policy  $\pi_\theta$ ) networks. Within each dense layer, the forward pass is split into two parallel paths:
\begin{enumerate}
    \item The Frozen Path (Blue): The input is processed by the pre-trained weights ($W_0$), ensuring the agent retains the foundational motor skills acquired in simulation.
    \item The Adapter Path (Green): The input is processed simultaneously by the trainable low-rank matrices ($A$ and $B$). As shown in the layer nodes, the adapter contribution is computed as $\frac{\alpha}{r} B A x$ where $\alpha$ >0 is a hyperparameter.
\end{enumerate}
These outputs are summed prior to the ELU activation function. For the Actor network specifically, a trainable noise parameter is injected at the final output layer to regulate exploration variance during the fine-tuning process.

Despite applying adapters to all layers, the low-rank constraint (where $\rho \ll d$) results in a dramatic reduction in computational overhead. This configuration reduces the number of trainable parameters by approximately $99.09\%$ over all tasks compared to \ac{FFT}, significantly accelerating the optimization landscape traversal.

We observe that the low-rank structure of LoRA acts as an intrinsic safety filter. By limiting the degrees of freedom available for adaptation, LoRA prevents the policy from overfitting to high-frequency noise or learning dangerous, bang-bang control, which causes spikes and is often associated with over-parameterized fine-tuning.

In later sections, we will empirically validate that a rank of $\rho=1$ is sufficient to capture the dominant correction direction required to align simulation-trained priors with physical hardware. This suggests that while simulation models may have unmodeled latencies, the foundational motor skills are largely correct; the reality gap is thus a linear manifold alignment problem rather than a task-relearning problem. 

\subsection{Safety Mechanisms} \label{sec:safety_mechanisms}


To prevent hardware damage during adaptation, we incorporate a discrete safety trigger that serves as a final decision gate. The filter monitors the robot's state $s_t$ at every timestep. If the filter predicts a violation of safety constraints (e.g., pitch/roll exceeding safe limits), it overrides the main policy and activates the {\em recovery policy} $\pi_r$. This mechanism ensures that the robot only explores safe states while the LoRA adapters refine the main policy. 

The recovery policy drives the system from a broad distribution of recoverable states back into a nominal safe state $s_{nom}$, rather than tracking a predefined recovery trajectory. In practice, $s_{nom}$ corresponds to a dynamically stable upright configuration with bounded base orientation and low velocity (e.g., standing), defined using only proprioceptive signals. This definition is independent of the downstream task and does not rely on robot-specific heuristics.

To ensure real-world robustness,  $\pi_r$ is trained using extensive domain randomization, including variations in mass, friction coefficients, and external forces. Training episodes are initialized across the robot’s operational envelope, and the reward encourages convergence toward $s_{nom}$ while assigning a large negative penalty to terminations, thereby strictly penalizing leaving the safe set. As a result, the policy learns to reliably return the system to a safe configuration regardless of the specific task being executed.

Since the primary objective is hardware safety rather than performance, the recovery policy is intentionally conservative, prioritizing successful recovery and avoiding mechanical failure. 
Crucially, because $\pi_r$ is task-agnostic and robustly trained, a single recovery controller suffices for multiple downstream tasks. 

\section{Experimental Results}
\label{sec:results}
In this section, we evaluate the performance, robustness, and safety of \name for fine-tuning different locomotion policies. Specifically, our experiments are structured to investigate:
\begin{itemize}
    \item \textbf{Safety and Robustness:} The reduction of mechanical failure rates during the policy fine-tuning phase.
    \item \textbf{Sample Efficiency:} The impact of low-rank parameterization on minimizing the required samples for transition from simulation to deployment-ready behaviors.
    \item \textbf{Generalization and Transfer:} The framework's performance across diverse locomotion tasks and its empirical validity on physical hardware (sim-to-real).
    \item \textbf{Ablation Analysis:} The degree to which performance gains are uniquely attributable to the \name architecture versus standard Proximal Policy Optimization \ac{PPO} hyperparameter tuning.
\end{itemize}

\subsection{Experimental Setup}

\begin{figure}
\centering
\includegraphics[width=0.7\linewidth]{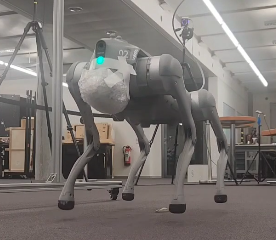}
\caption{The Unitree Go2 robot performs a dynamic jump. Using \name, we fine-tune a low-rank adapter on top of a frozen simulation-trained policy, achieving substantially faster training with near-zero safety violations compared to standard \ac{PPO}.}
\label{fig:2}
\vspace{-0.25cm}
\end{figure}

We validate our framework on the Unitree Go2 quadruped robot, a platform characterized by high-torque actuators and dynamic agility, as shown in Figure \ref{fig:2}. To assess the versatility of our framework, we evaluate two distinct locomotion behaviors: trotting, which tests endurance and cyclic stability, and jumping, a highly dynamic task requiring precise impulse coordination.

We benchmark \name against three established paradigms:
\begin{itemize} 
    \item \textbf{\ac{PPO} (Zero-Shot):} Represents the direct deployment of the robust simulation-trained policy without fine-tuning. 
    \item \textbf{Full Fine-Tuning (FFT) \underline{w}ith\underline{o}ut safety:} Represents the naive implementation of conventional full-parameter updates directly on the hardware. This baseline excludes safety filters and recovery policies to quantify the mechanical-failure risks inherent in unconstrained exploration. 
    \item \textbf{Full Fine-Tuning (FFT) \underline{w}ith safety:} Augments the standard full-parameter fine-tuning with our safety filter and recovery policy. This serves as a strong baseline to isolate the sample efficiency gains of \ac{LoRA} from the benefits of the safety mechanism alone. 
\end{itemize}
To ensure a rigorous comparison, all methods utilize identical \ac{PPO} hyperparameters and reward functions. The learning rates were optimized individually, $10^{-2}$ for \ac{LoRA} and $10^{-3}$ for FFT, to allow each algorithm to achieve its peak convergence speed. We evaluate performance based on cumulative reward, wall-clock convergence time, and the frequency of safety violations.

Fine-tuning a policy directly on robotic hardware could be dangerous, since the policy may temporarily or permanently degrade during refinement. Therefore, we develop two types of validation protocols:
\begin{itemize} 
\item \textbf{Sim-to-Sim Protocol}: To test distribution shift handling, we use a sim-to-sim transfer protocol in which the base policy is pre-trained in IsaacSim \citep{NVIDIA_Isaac_Sim} (PhysX engine) and adapted to MuJoCo \citep{todorov2012mujoco}, running in real-time just like it would on the real robot. This sim-to-sim transfer has physics mismatch due to differences in contact solvers and integrators, as well as the constraints imposed by real-time execution on physical robots.

\item \textbf{Sim-to-Real Protocol:}
For real-world experiments, we use a Vicon motion-capture system to provide ground-truth base position and velocity. To account for stochasticity, all experiments are averaged across 4 random seeds. Crucially, fine-tuning runs are executed in real time on a workstation (20-core CPU, RTX 3080 Ti), ensuring that the control loop maintains the required frequency with minimal computational bottlenecks.

\end{itemize}
\subsection{Safety During Fine-tuning}

In this section, we evaluate the safety of \name against baseline methods by recording mechanical failures (e.g., falls, collisions, or joint limit violations) during the training period. Table \ref{tab:safety} summarizes the mechanical failures recorded during fine-tuning.


In this setting, the pre-trained policy is intentionally quite fragile, making it highly sensitive to aggressive gradient updates. As demonstrated in Table \ref{tab:safety}, \ac{FFT} without safety frequently failed to recover, with an average of $14.25$ failures for trot and $x$ for jump forward. Even with safety mechanisms, \ac{FFT} still averaged $7.5$ failures for trot and $27$ for jump. In contrast, \name maintained a zero-failure rate across all seeds. These results suggest that operating within a low-rank adaptation space effectively filters out destabilizing gradient directions, enabling \name to function as an implicit recovery policy during fine-tuning.

\begin{table}[h]
\centering
\caption{\textbf{Safety Evaluation:} Average failures (falls/crashes) during the training across 4 seeds.\label{tab:safety}}
\small 
\begin{tabular}{lcc}
\toprule
\textbf{Task} & \textbf{Trot} & \textbf{Jump} 
\\
\midrule

\makecell[l]{\ac{FFT}  \textit{wo safety}} & $ 14.25$      & $69.0$             \\
\addlinespace

\makecell[l]{\ac{FFT} \textit{w safety}} & $ 7.5$      & $17.5$             \\
\addlinespace

\textbf{\textit{}{SLowRL}(Ours)}    & $\mathbf{0.0}$  & $\mathbf{2.0}$   \\
\bottomrule
\end{tabular}
\vspace{-0.25cm}
\end{table}

In addition to discrete failure counts, we evaluate control smoothness during training using the action rate, defined as $\|a_t - a_{t-1}\|$. This metric directly reflects the presence of high-frequency, bang-bang control behavior discussed in Sec.~\ref{LoRA_PPO}. 

As shown in Table \ref{tab:action_rate}, \name consistently maintains a significantly lower action rate during fine-tuning. For the trot gait, \name achieved a $88.9\%$ reduction in action rate—a $50\%$ improvement over the \ac{FFT} baseline. Even in the more dynamic Jump gait, where \ac{FFT} only reduced the action rate by $5.7\%$, \name achieved a $35.7\%$ reduction. This indicates that the low-rank constraint effectively suppresses unsafe high-frequency control updates in practice.
\begin{table}[h]
\centering
\caption{\textbf{Smoothness:} Percentage reduction in action rate $\|a_t - a_{t-1}\|$ from start to finish. \label{tab:action_rate}}
\small 
\setlength{\tabcolsep}{4pt}
\begin{tabular}{lcc}
\toprule
\textbf{Gait} & \makecell{FFT (Reduction \%)} & \makecell{\name(Reduction \%)} 
\\
\midrule
\makecell[l]{Trot} & 38.9\%      & 88.9\%            \\
\addlinespace
\makecell[l]{Jump} & 5.7\%     &35.7\%           \\
\bottomrule
\end{tabular}
\vspace{-0.25cm}
\end{table}


\subsection{Sample Efficiency during Fine-tuning} 
\begin{figure}[t]
\centering
\includegraphics[width=\linewidth]{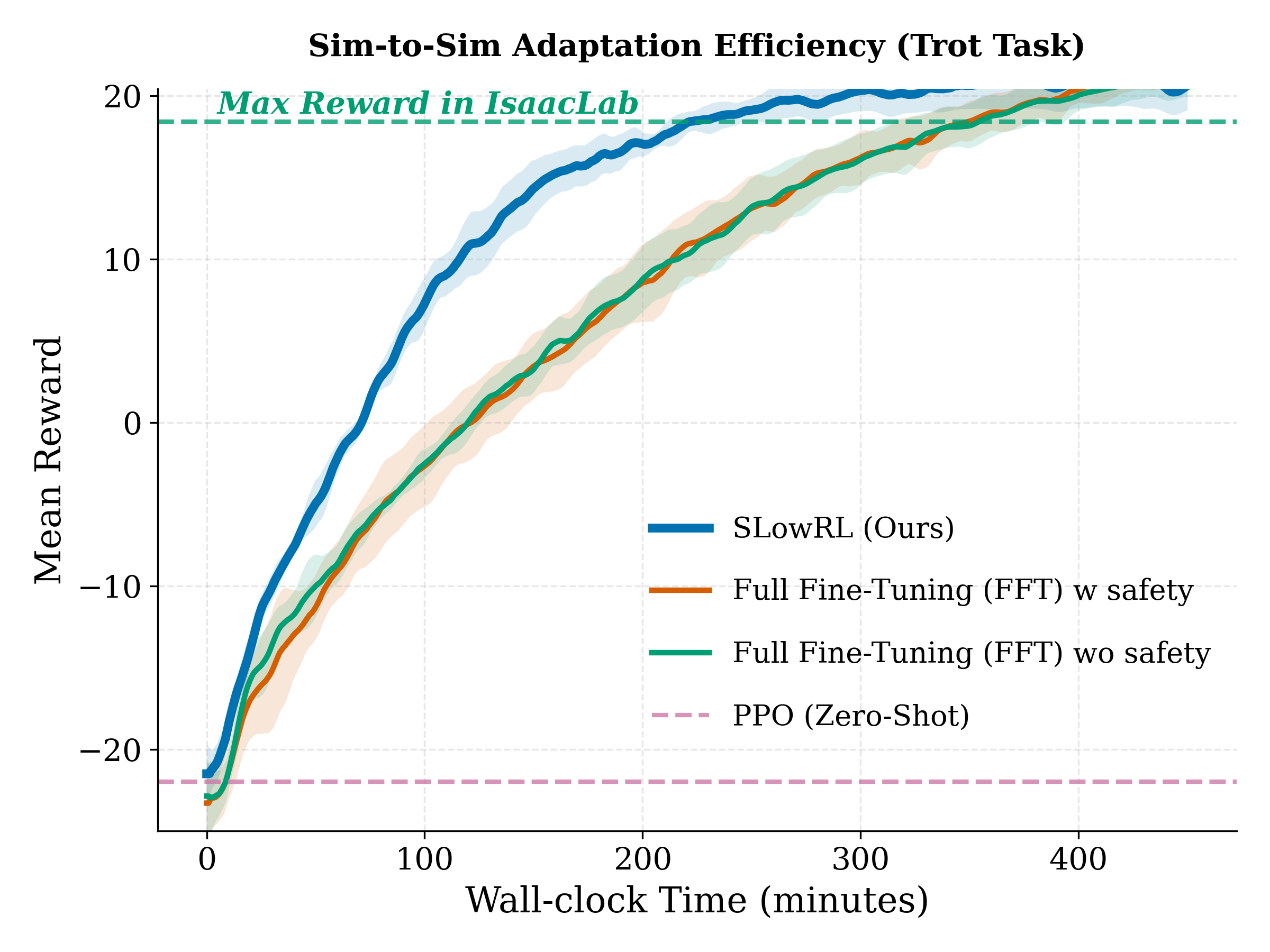}
\caption{\textbf{Sim-to-Sim Sample Efficiency during Fine-tuning (Trot Task)} Comparative learning curves showing mean reward over wall-clock time for the trotting task. \name (blue) significantly outperforms the \ac{FFT} and Zero-Shot baselines, achieving a $38\%$ reduction in time-to-convergence while maintaining higher stability.}
\label{fig:trot_curves}
\vspace{-0.45cm}
\end{figure}

\begin{figure}[t]
\centering
\includegraphics[width=\linewidth]{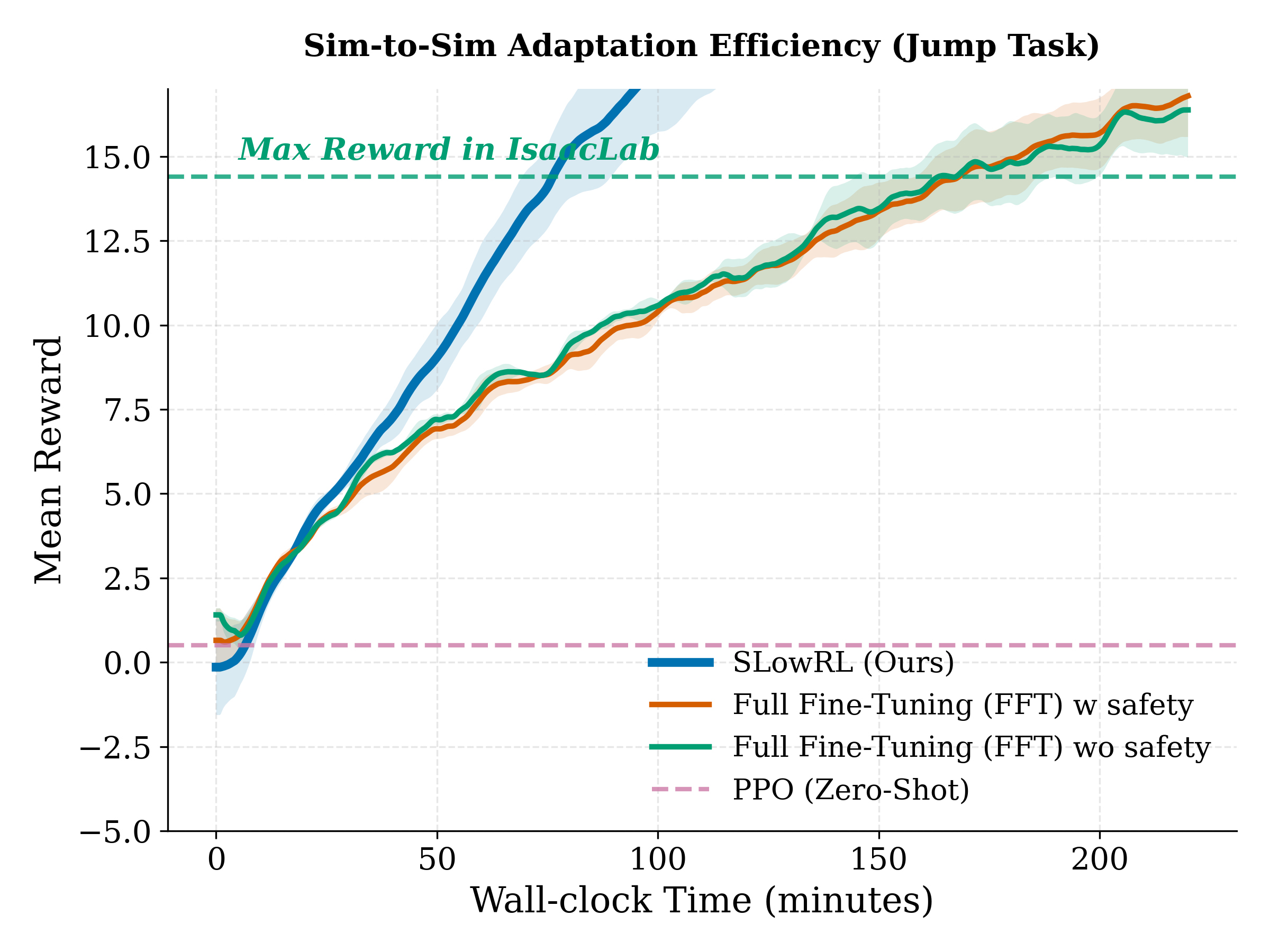}
\caption{\textbf{Sim-to-Sim Sample Efficiency during Fine-tuning (Jump Task)} Performance comparison for the dynamic jumping task on the Unitree Go2 robot. \name demonstrates superior sample efficiency, reducing convergence time by $55\%$ compared to baselines without incurring the safety violations observed in \ac{FFT}.}
\label{fig:jump_curve}
\vspace{-0.45cm}
\end{figure}

We use a sim-to-sim (IsaacLab to MuJoCo) setting to quantify the sample efficiency of \name. Here, we use wall-clock convergence time as a proxy for sample efficiency during fine-tuning. Faster convergence directly corresponds to fewer physical interactions and reduced mechanical risk. Figures \ref{fig:trot_curves} and \ref{fig:jump_curve} show the results for trot and jump.
Quantitatively, \name reduces the time required to match the original IsaacLab performance to $38\%$ for trotting and $55\%$ for jumping. Finally, because the failure count is low (under 20), \ac{FFT} with and without safety mechanisms exhibits a nearly identical pattern of learning.

\subsection{The Effect of Rank ($\rho$)}
In Section~\ref{LoRA_PPO}, we argued that low-rank updates may be sufficient for learning under idealized assumptions. Whether such extreme low-rank adaptation remains effective in contact-rich, safety-critical robotic control, however, remains an open empirical question.

To investigate this, we sweep the \ac{LoRA} rank parameter $\rho \in \{1, 2, 4, 8\}$ while keeping all other hyperparameters fixed. All configurations are trained for the same fixed wall-clock duration of 75 minutes. This time horizon corresponds to the point at which the rank-1 configuration reaches the maximum reward achieved by the pre-trained simulation policy. Importantly, this budget is applied uniformly to all ranks. To further verify that this choice does not bias the comparison, we additionally ran all rank configurations for an extended duration of 230 minutes; across this longer horizon, lower-rank adapters consistently achieved the highest rewards in average, and higher-rank configurations did not surpass rank-1 performance.

\begin{figure}[t]
\centering
\includegraphics[width=\linewidth]{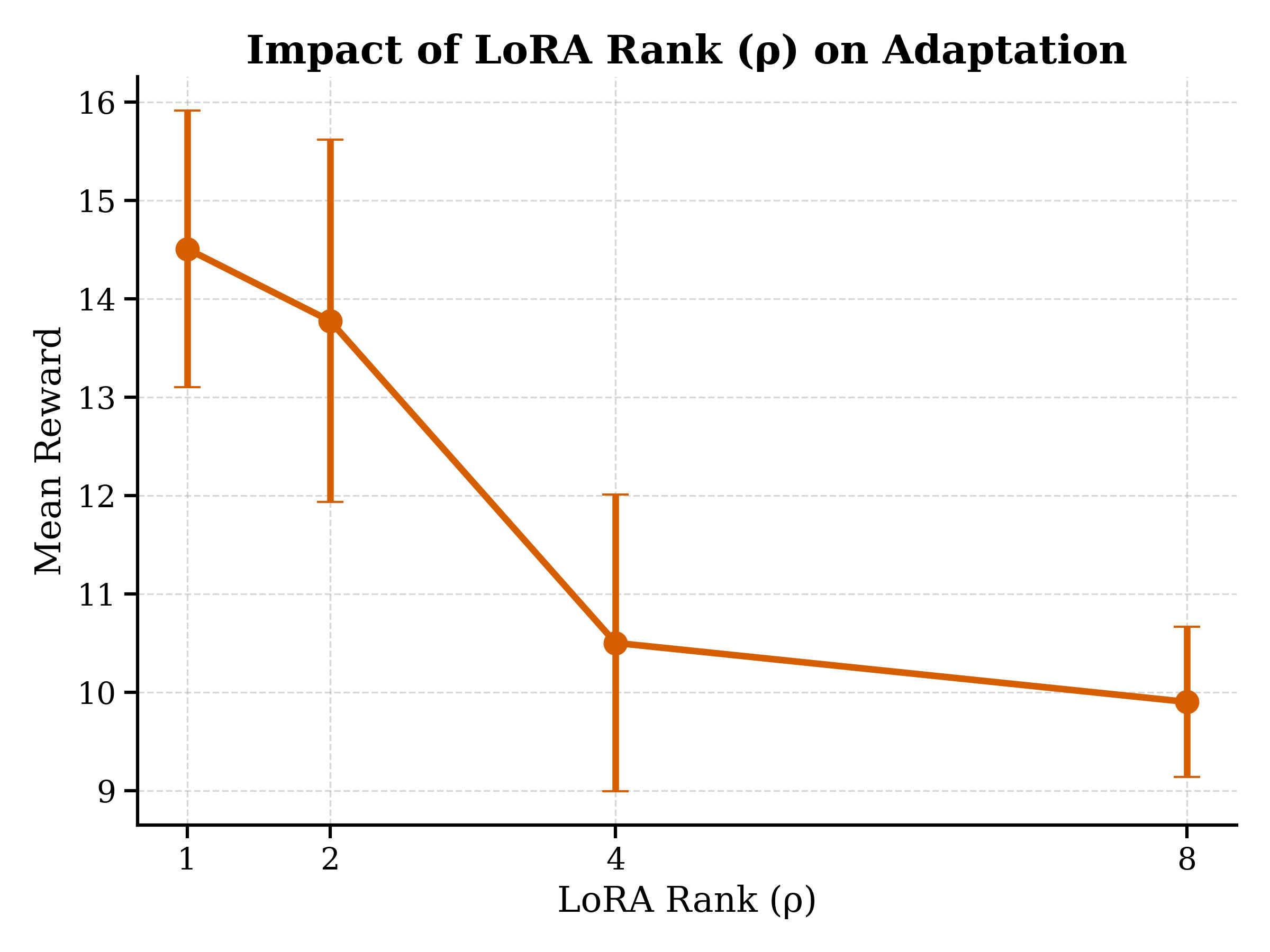}
\caption{\textbf{Impact of LoRA Rank ($\rho$) on Adaptation} An ablation study sweeping rank parameters $\rho \in \{1, 2, 4, 8\}$. The results empirically validate that a minimal rank of $\rho=1$ yields the highest final reward, confirming that effective sim-to-real adaptation can be achieved via a single principal direction of adjustment.}
\label{fig:rank}
\vspace{-0.45cm}
\end{figure}

Figure \ref{fig:rank} reports the resulting learning curves under this fixed-budget protocol. We observe that rank-1 adaptation $(\rho=1)$ reaches the pre-trained performance level fastest and within the allotted time, while higher-rank configurations converge more slowly and, in some cases, fail to reach the same performance threshold within the same time budget.

These results indicate that increasing the rank does not improve adaptation efficiency under fixed training budgets. While higher-rank adapters introduce greater expressive capacity, they also expand the dimensionality of the optimization problem, amplifying gradient noise in the presence of contact discontinuities, delayed rewards, and safety-triggered resets.

From a practical perspective, this suggests that the dominant mismatch between pre-trained and target dynamics can be corrected within a low-dimensional subspace for a fixed task. Rank-1 adaptation captures this dominant correction direction most efficiently, whereas additional degrees of freedom primarily delay convergence without improving final performance. Consequently, all subsequent experiments adopt rank-1 \ac{LoRA} adapters.

\subsection{Ablation Analysis} \label{Lora_structure}
To determine the optimal configuration for using \ac{LoRA} in a quadruped robotics setting to improve sample efficiency for online adaptation, we investigate multiple locations at which \ac{LoRA} can be added to the policy architecture. Our study evaluates the placement of \ac{LoRA} modules across two primary dimensions: the functional role of the sub-networks (actor vs. critic) and the specific depth of the internal layers being adapted.

\subsubsection{Architectural Components, Actor vs. Critic}
We first compared the necessity of adapting the value function versus the policy by evaluating configurations where adapters were applied exclusively to the Actor, and where they were applied to both the Actor and Critic.

\begin{figure}[t]
\centering
\includegraphics[width=\linewidth]{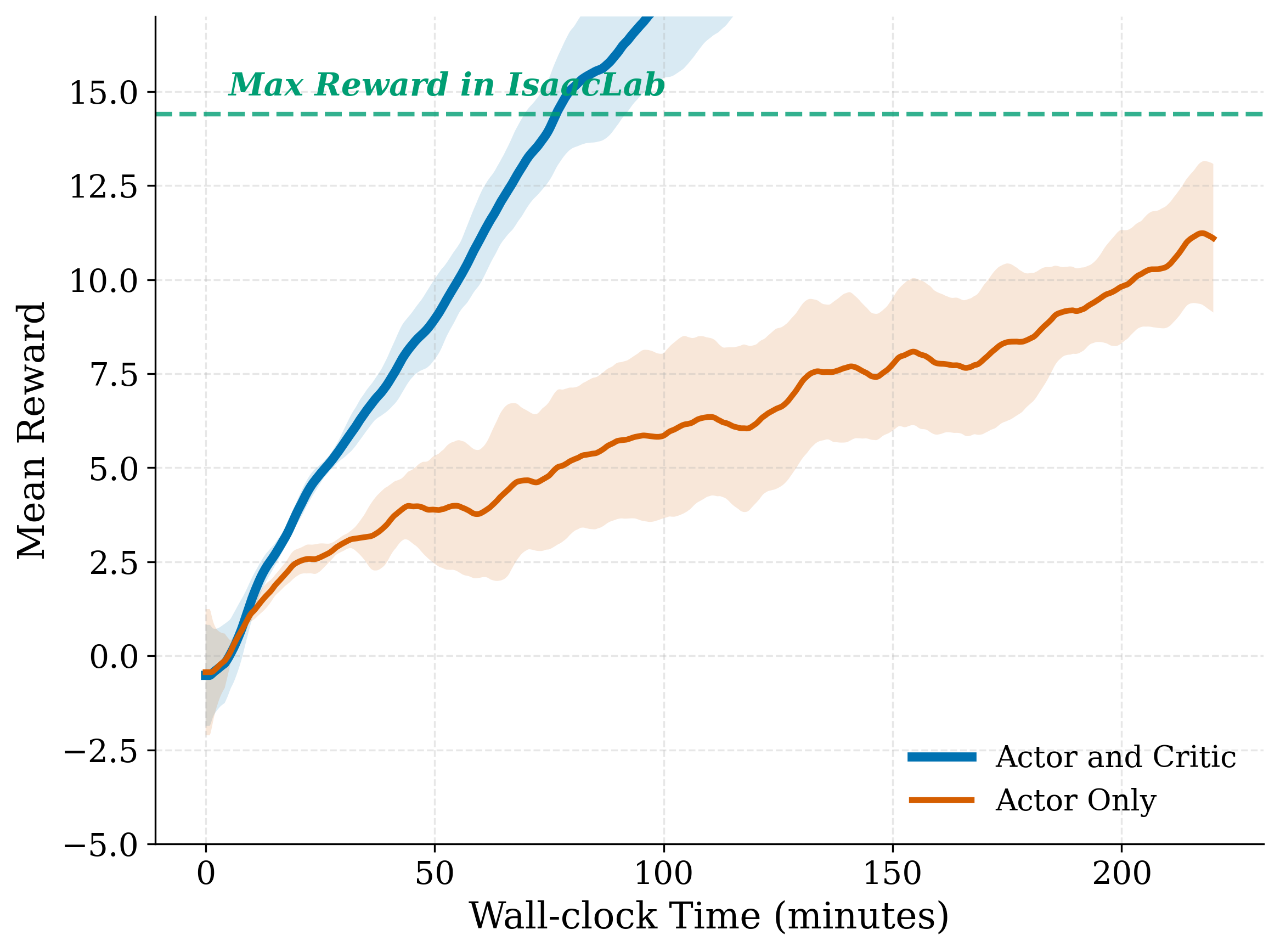}
\caption{\textbf{Impact of Critic Adaptation} A comparison of adaptation performance when applying \ac{LoRA} to the Actor and Critic versus the Actor Only. The results show that adapting the critic is strictly necessary to realign the value function with real-world dynamics, whereas the Actor Only configuration fails to converge.}
\label{fig:actor_vs_both}
\vspace{-0.45cm}
\end{figure}

As illustrated in Figure \ref{fig:actor_vs_both}, our experiments demonstrate that applying \ac{LoRA} to both the Actor and Critic yields the highest performance and stability. Notably, the Actor Only configuration, where the Critic remains frozen at its pre-trained, it fails to converge to an optimal gait. We attribute this to the significant distribution shift between the source (IsaacLab) and target (MuJoCo/Real) environments. A frozen Critic continues to evaluate states based on simulation physics, providing inaccurate advantage estimates that destabilize the Actor's adaptation. 

Therefore, adapting the critic is strictly necessary to realign the value function with the physical reality of the target domain. This necessity is further evidenced by the final value-function loss recorded at the end of the fine-tuning phase. While the Actor and Critic configuration successfully minimizes the loss to a highly precise level of $0.007$, the Actor Only setup maintains a substantial residual loss of $0.2$.

This two-order-of-magnitude difference in error underscores the indispensable value of critic adaptation for sim-to-real transfer. In the Actor Only configuration, the critic remains anchored to the source environment's dynamics, resulting in a persistent and severe estimation bias. Without fine-tuning, the pre-trained value function becomes fundamentally misaligned with the target domain's reward manifold.

Specifically, the frozen critic remains anchored to the source dynamics, resulting in a fundamental misalignment between the predicted state-values and the target domain’s reward manifold. This creates a divergent value baseline that provides the Actor with deceptive advantage estimates. The $0.2$ loss plateau demonstrates that without recalibrating the critic, the agent’s internal performance metric is inaccurate, rendering stable policy refinement in a new domain mathematically untenable.

\subsubsection{Structural Depth, Layer-wise Injection}
We further analyzed the impact of LoRA placement across network depths. We evaluated four configurations:
\begin{itemize}
    \item Output Layer Only: Adapting only the final action projection.
    \item Input \& Output: Adapting the first feature encoder layer and the final projection.
    \item All Layers: Injecting adapters into every dense layer of the policy.
\end{itemize}

\begin{figure}[t]
\centering
\includegraphics[width=\linewidth]{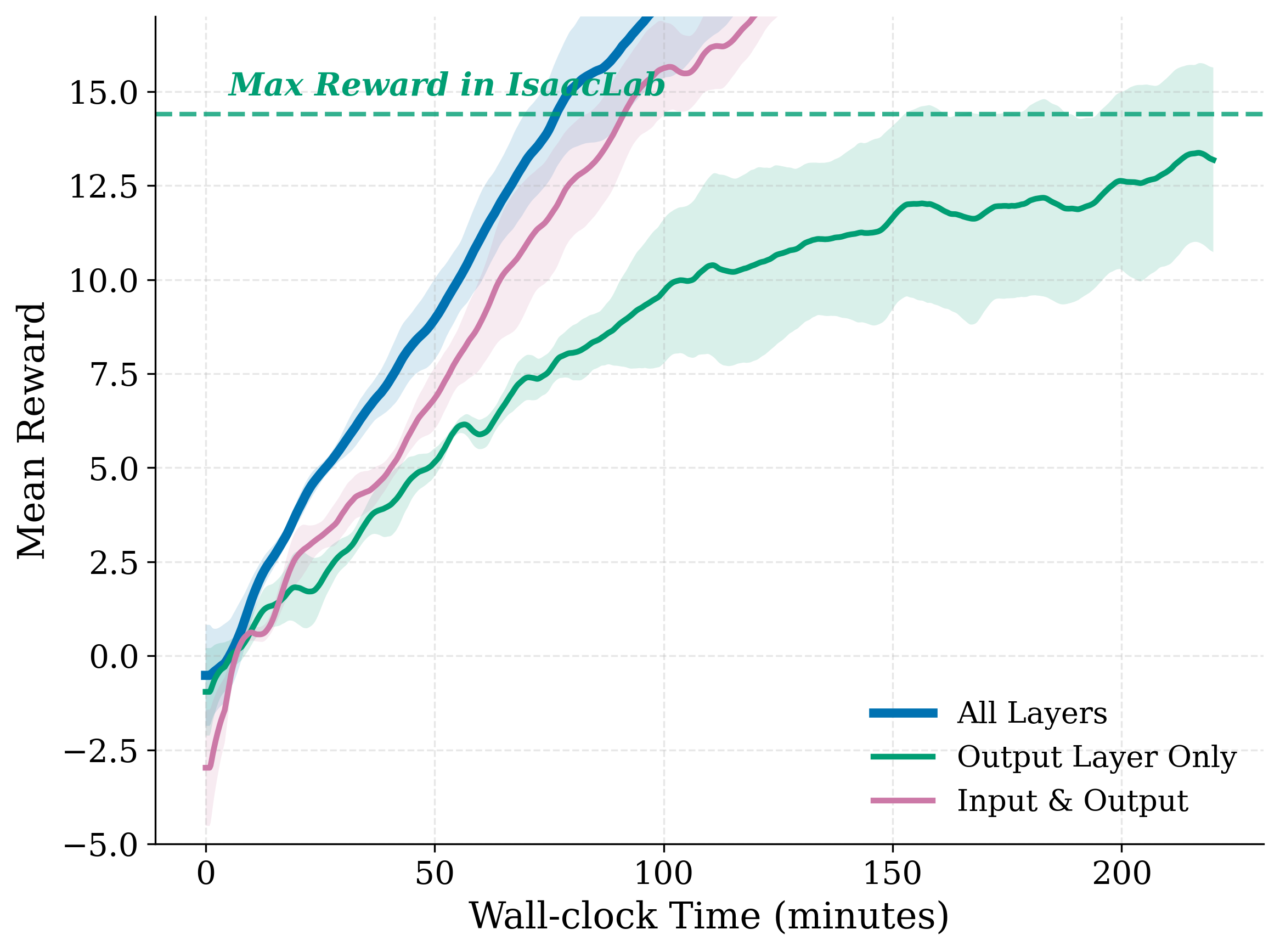}
\caption{\textbf{Evaluation of Adapter Placement Strategy} Performance comparison of different \ac{LoRA} injection depths, including partial layer adaptation. The All Layers configuration achieves the highest performance, demonstrating that deep correction across all levels of abstraction is required to robustly bridge the sim-to-real gap.}
\label{fig:layers}
\vspace{-0.45cm}
\end{figure}

Our empirical results in Figure \ref{fig:layers} show that applying LoRA to All Layers yields the highest performance, surpassing the parallel residual adapter and partial-network methods.

While the parallel residual adapter intuitively allows for direct action correction, we found it struggled to compensate for internal feature mismatches. By contrast, the All Layers configuration allows the policy to perform deep correction and adjust how the robot processes state features at every level of abstraction, resulting in the most robust transfer.

\subsection{Real-World experiments}
\begin{figure}[t]
\centering
\includegraphics[width=\linewidth]{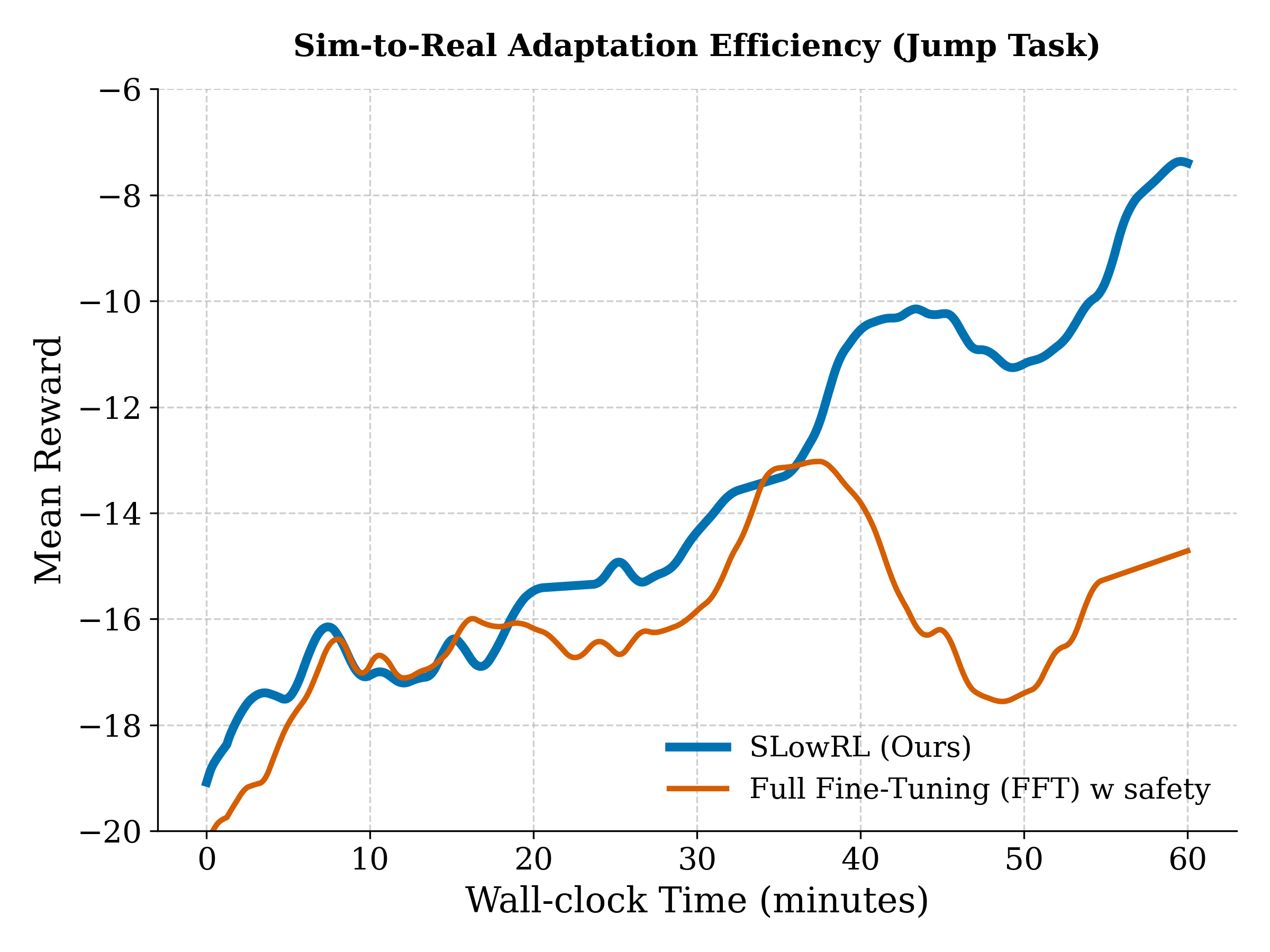}
\caption{\textbf{Sim-to-Real Adaptation Efficiency (Jump Task)} A comparison of mean reward over wall-clock time. \name (blue) demonstrates faster convergence and higher asymptotic performance compared to \ac{FFT} with safety constraints (orange).}
\label{fig:real_jump_curves}
\vspace{-0.45cm}
\end{figure}

To demonstrate the effectiveness of \name for fine-tuning the jump task in the real world, we compared its performance with \ac{FFT} in Figure~\ref{fig:real_jump_curves}.
Quantitatively, the \name agent demonstrates superior adaptation compared to baselines. As illustrated in Figure~\ref{fig:real_jump_curves}, \name rapidly improves the mean reward from $-19$ to $-7$ in 60 minutes, significantly outperforming \ac{FFT}, which plateaus near $-15$. Regarding gait quality, the agent achieves a 9\% improvement in contact pattern consistency for the jumping task.

\section{Discussion}
\label{sec:discussion}
In this work, we presented \name, a unified framework for safe and efficient adaptation of locomotion policies from simulation to the real world. Our findings challenge the prevailing assumption in robotic reinforcement learning that sim-to-real transfer requires either massive domain randomization or computationally expensive full-parameter fine-tuning. By dissecting the fine-tuning process, we offer several key insights into the nature of the reality gap and the mechanisms required to bridge it.

The Low-Rank Hypothesis of Sim-to-Real Transfer:
Our most significant finding is the efficacy of extremely low-rank updates ($\rho=1$) in bridging the sim-to-real gap. The success of minimal rank adaptation suggests that the foundational motor skills learned in high-fidelity simulation ($W_0$) are largely correct; the reality gap does not necessitate a fundamental relearning of locomotion primitives, but rather a linear realignment of the policy's manifold. This aligns with recent theoretical work in \ac{LoRA} without regret \cite{schulman2025lora}, indicating that in the low-information regime of real-world \ac{RL}, the effective dimensionality of the optimal policy update is small. By constraining the optimization to this low-rank subspace, \name implicitly regularizes the policy, preventing the forgetting of safe priors and the emergence of high-frequency, bang-bang control artifacts often seen in \ac{FFT}.

The Necessity of Value Re-Alignment:
Our ablation studies reveal a critical, often overlooked component of fine-tuning: the adaptation of the critic. While previous work has sometimes focused exclusively on policy (Actor) adaptation, our results demonstrate that freezing the critic leads to convergence failure. The significant residual loss (0.2) in Actor Only configurations confirms that the perception of value is as susceptible to the sim-to-real gap as the control policy itself. When the critic remains anchored to simulation dynamics, it provides deceptive advantage estimates that destabilize learning. Therefore, we posit that robust sim-to-real transfer requires a synchronized realignment of both the actor's control strategy and the critic's value estimation.

Safety as an Enabler of Efficiency:
A central insight of our work is that safety constraints function as an enabler of learning efficiency rather than a hindrance. By offloading the risk of catastrophic failure to a task-agnostic Recovery Policy, the learning agent is free to explore the local solution space more aggressively. \name achieved a $46.5\%$ reduction in convergence time precisely because it eliminated the conservative restarts and mechanical downtime associated with the frequent falls observed in \ac{FFT}. The zero-failure rate observed during training stands in stark contrast to the frequent falls in \ac{FFT}, validating the architecture's ability to decouple exploration from structural integrity.

\section{Limitations and Future Work}
Although \name demonstrates robust performance in dynamic gaits such as trotting and jumping, our current evaluation is limited to flat ground and moderate disturbances. Adapting to highly unstructured terrains (e.g., stairs, debris) may require a higher-rank adaptation ($\rho > 1$) to capture more complex contact interactions. 

A significant limitation of the current Recovery Policy is its reliance on a human-designed safe region. Our results indicate that attempting to learn this safety manifold purely in simulation is often ineffective due to a significant critic mismatch; the boundaries optimized by the agent in a virtual environment do not accurately reflect the physical realities and contact dynamics of the real world. To address this, future work will focus on developing methods to learn these safety regions directly on physical hardware autonomously.

Furthermore, reliance on \ac{PPO} presents a significant bottleneck for real-world deployment. As an on-policy algorithm, PPO is notoriously sample-inefficient and lacks the mechanism to utilize previous data collected during real-world interactions.
Consequently, we plan to investigate off-policy algorithms and hybrid fine-tuning architectures that can leverage historical transitions and large-scale data to achieve faster, more robust adaptation than is currently possible with standard on-policy methods.

\section{Conclusion}
\label{sec:Conclusion}
This paper introduced \name, a parameter-efficient fine-tuning framework that resolves the safety paradox of real-world robotic learning: the tension between the need for physical interaction to improve performance and the risk that such interaction poses to hardware. By combining \ac{LoRA} with a robust separate safety filter, we demonstrated that it is possible to adapt simulation-trained policies to physical robots rapidly and safely.

Our extensive empirical evaluation on the Unitree Go2 quadruped validates that full-model retraining is unnecessary for handling sim-to-real dynamics mismatches. Instead, optimizing a low-rank subspace enables an adaptation that is, on average, $46.5\%$ faster than standard PPO baselines, thus eliminating safety violations. We further established that this efficiency relies on the simultaneous adaptation of both Actor and Critic to strictly realign the agent's value expectations with physical reality.

Ultimately, \name provides a practical blueprint for the last mile of robotic deployment. As foundation models and large-scale simulations become the standard for pre-training, lightweight, safe, and efficient adaptation methods like \name will be essential for grounding these generalist policies in the specific, unmodeled realities of the physical world. This work represents a significant step toward robots that can continuously adapt themselves to their environment without the intervention of engineers or the risk of self-destruction.

\bibliographystyle{unsrt}
\bibliography{references} 
\clearpage

\end{document}




\setlength{\abovecaptionskip}{3pt}
\setlength{\belowcaptionskip}{0pt}

\section{appendix}
\section*{Appendix A: Low-Rank Structure of Sim-to-Real Adaptation}

In this appendix, we formalize the relationship between low-rank adaptation and sim-to-real transfer. The results do not assume that the sim-to-real gap is inherently rank-1; instead, they establish conditions under which low-rank adaptation is sufficient or optimal.

\subsection*{A.1 Exact Expressivity of Low-Rank Adaptation}

\begin{proposition}[Sufficiency of LoRA for Low-Rank Adaptation]
Let $W_0 \in \mathbb{R}^{d \times k}$ denote the frozen weights of a linear layer, and suppose the optimal real-world weights are given by
\[
W^\star = W_0 + \Delta W,
\]
where $\operatorname{rank}(\Delta W) = r$. Then there exist matrices $B \in \mathbb{R}^{d \times r}$ and $A \in \mathbb{R}^{r \times k}$ such that
\[
\Delta W = BA.
\]
Consequently, LoRA with rank $\rho \ge r$ can represent the optimal adapted weights exactly.
\end{proposition}

\begin{proof}
Any matrix $\Delta W$ of rank $r$ admits a rank factorization $\Delta W = UV^\top$, where $U \in \mathbb{R}^{d \times r}$ and $V \in \mathbb{R}^{k \times r}$ have full column rank. Setting $B := U$ and $A := V^\top$ yields $\Delta W = BA$. \qedhere
\end{proof}

\paragraph{Interpretation.}
If the true sim-to-real correction required at a layer is low-rank, then low-rank adaptation is not an approximation but an exact parameterization.

---

\subsection*{A.2 Optimality of Rank-1 Adaptation}

In practice, $\Delta W$ need not be exactly rank-1. However, rank-1 adaptation remains optimal among all rank-1 updates in a precise sense.

\begin{proposition}[Optimal Rank-1 Approximation]
Let $\Delta W \in \mathbb{R}^{d \times k}$ be any matrix with singular value decomposition
\[
\Delta W = \sum_{i=1}^{\min(d,k)} \sigma_i u_i v_i^\top,
\quad \sigma_1 \ge \sigma_2 \ge \dots \ge 0.
\]
Then the solution to
\[
\arg\min_{\operatorname{rank}(\widehat{\Delta W}) \le 1}
\|\Delta W - \widehat{\Delta W}\|_F
\]
is given by
\[
\widehat{\Delta W} = \sigma_1 u_1 v_1^\top.
\]
\end{proposition}

\begin{proof}
This follows directly from the Eckart--Young--Mirsky theorem, which states that truncating the singular value decomposition yields the best rank-$\rho$ approximation in Frobenius norm. \qedhere
\end{proof}

\paragraph{Interpretation.}
When constrained to rank-1 updates, LoRA recovers the best possible approximation to the true parameter correction in Frobenius norm. Empirical success of rank-1 adaptation therefore implies that the dominant singular mode of $\Delta W$ captures most of the sim-to-real discrepancy.

---

\subsection*{A.3 Low-Dimensional Reality Gap Implies Low-Dimensional Parameter Updates}

We now provide conditions under which low-rank parameter corrections are expected.

\begin{proposition}[Low-Dimensional Reality Gap]
Let $p \in \mathbb{R}^m$ denote physical environment parameters (e.g., friction coefficients, latency, mass distribution), and let $\theta(p)$ denote the optimal policy parameters under dynamics $p$. Assume $\theta(p)$ is differentiable in a neighborhood of the simulator parameters $p_0$. Then for small mismatch $\delta p = p^\star - p_0$,
\[
\theta(p^\star) - \theta(p_0) \approx J_\theta(p_0)\, \delta p,
\]
where $J_\theta(p_0) \in \mathbb{R}^{|\theta| \times m}$ is the Jacobian.
\end{proposition}

\begin{proof}
The result follows from a first-order Taylor expansion of $\theta(p)$ around $p_0$. \qedhere
\end{proof}

\paragraph{Interpretation.}
The parameter update lies in the span of at most $m$ directions. When the real-world differs from simulation along a small number of physical axes, the resulting policy update is confined to a low-dimensional subspace. If these directions induce correlated structure across layers, the induced weight updates can be well-approximated by low-rank matrices.

---

\subsection*{A.4 Implications for Sim-to-Real Fine-Tuning}

Together, these results show that:
\begin{itemize}
\item If the true sim-to-real correction is low-rank, LoRA of sufficient rank is exactly expressive.
\item If constrained to rank-1 adaptation, LoRA yields the optimal rank-1 correction.
\item When the reality gap is governed by a small number of physical parameters and the optimal policy varies smoothly with them, low-rank parameter updates are expected.
\end{itemize}

These theoretical observations provide a principled explanation for the empirical effectiveness of rank-1 adaptation observed in our experiments.
